\documentclass[conference]{IEEEtran}
\IEEEoverridecommandlockouts
\usepackage{cite}
\usepackage{amsmath,amssymb,amsfonts}
\usepackage{algorithmic}
\usepackage{graphicx}
\graphicspath{{figures/}}
\usepackage{textcomp}
\usepackage{xcolor}

\usepackage[utf8]{inputenc}
\usepackage[hyphens]{url}
\usepackage{cleveref}
\usepackage{xspace}
\usepackage{subfigure}

\IEEEpubid{\begin{minipage}[t][10cm][t]{\textwidth}\ \\[10pt] \centering
		{\normalsize\textcopyright 2019 IEEE. Personal use of this material is permitted.  Permission from IEEE must be obtained for all other uses, in any current or future media, including reprinting/republishing this material for advertising or promotional purposes, creating new collective works, for resale or redistribution to servers or lists, or reuse of any copyrighted component of this work in other works.}
	\end{minipage}}

\def\BibTeX{{\rm B\kern-.05em{\sc i\kern-.025em b}\kern-.08em
    T\kern-.1667em\lower.7ex\hbox{E}\kern-.125emX}}
\begin{document}

\title{
	Effortless Deep Training for Traffic Sign Detection Using Templates and Arbitrary Natural Images\\
	\thanks{This study was financed in part by Coordenação de Aperfeiçoamento de Pessoal de Nível Superior – Brasil (CAPES) – Finance Code 001; Conselho Nacional de Desenvolvimento Científico e Tecnológico – Brasil (CNPq) – grants 311120/2016-4 and 311504/2017-5 and of PIIC UFES (grant 138682/2018-6); and Fundação de Amparo à Pesquisa do Espírito Santo - Brazil (FAPES) – grant 84412844/2018.
	}
}

\author{
	\IEEEauthorblockN{Lucas Tabelini Torres\IEEEauthorrefmark{1}\IEEEauthorrefmark{4}, Thiago M. Paix\~ao\IEEEauthorrefmark{1}\IEEEauthorrefmark{2}, Rodrigo F. Berriel\IEEEauthorrefmark{1}, Alberto F. De Souza, \textit{Senior Member}, \textit{IEEE}\IEEEauthorrefmark{1},\\Claudine Badue\IEEEauthorrefmark{1}, Nicu Sebe\IEEEauthorrefmark{3} and Thiago Oliveira-Santos\IEEEauthorrefmark{1}} 
	\IEEEauthorblockA{\IEEEauthorrefmark{1}Universidade Federal do Esp\'irito Santo, Brazil}
	\IEEEauthorblockA{\IEEEauthorrefmark{2}Instituto Federal do Esp\'irito Santo, Brazil}
	\IEEEauthorblockA{\IEEEauthorrefmark{3}University of Trento, Italy}
	\IEEEauthorblockA{\IEEEauthorrefmark{4}Email: tabelini@lcad.inf.ufes.br}
}

\maketitle

\begin{abstract}
	Deep learning has been successfully applied to several problems related to autonomous driving. Often, these solutions rely on large networks that require databases of real image samples of the problem (i.e., real world) for proper training. The acquisition of such real-world data sets is not always possible in the autonomous driving context, and sometimes their annotation is not feasible (e.g., takes too long or is too expensive). Moreover, in many tasks, there is an intrinsic data imbalance that most learning-based methods struggle to cope with. It turns out that traffic sign detection is a problem in which these three issues are seen altogether. In this work, we propose a novel database generation method that requires only (i) arbitrary natural images, i.e., requires no real image from the domain of interest, and (ii) templates of the traffic signs, i.e., templates synthetically created to illustrate the appearance of the category of a traffic sign. The effortlessly generated training database is shown to be effective for the training of a deep detector (such as Faster R-CNN) on German traffic signs, achieving 95.66\% of mAP on average. In addition, the proposed method is able to detect traffic signs with an average precision, recall and F1-score of about 94\%, 91\% and 93\%, respectively. The experiments surprisingly show that detectors can be trained with simple data generation methods and without problem domain data for the background, which is in the opposite direction of the common sense for deep learning. 
\end{abstract}


\

\section{Introduction}

\begin{figure*}[t]
	\centering
	\includegraphics[width=\linewidth]{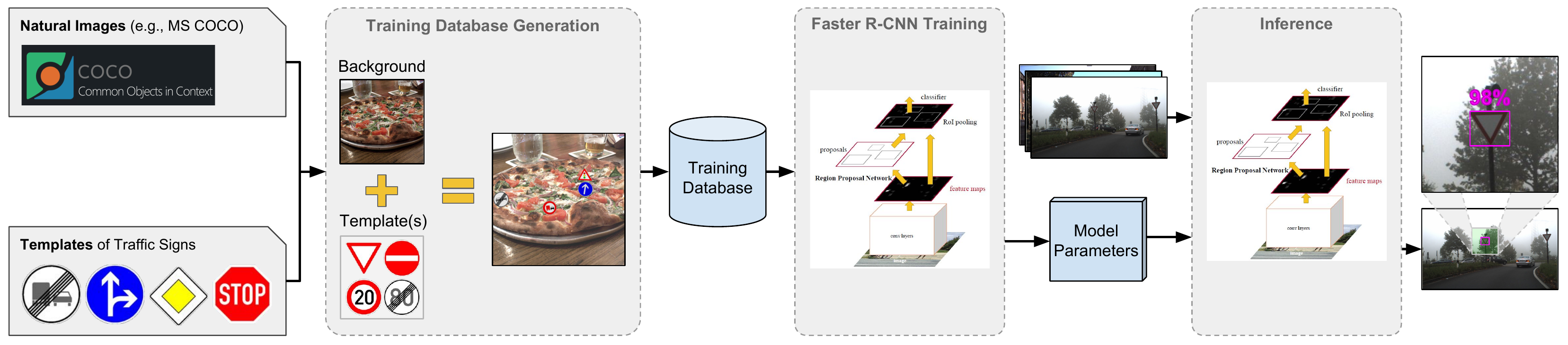}
	\caption{Overview of the proposal method. From left to right, the method receives as input natural images (e.g., from publicly available large-scale databases) and templates of traffic signs and generates a synthetic training database. The synthetic database is used to train a deep detector (e.g., Faster R-CNN). Finally, the model is ready to detect traffic signs.}
	\label{fig:overview}
\end{figure*}

Deep learning has been widely used to tackle a variety of computer vision tasks. Deep neural networks (DNNs) have also been successfully applied on several problems related to autonomous driving~\cite{oursurvey2019arxiv}. Many of these applications rely on large networks which in turn usually require large amounts of data to be properly trained. This requirement, however, is not always easy to be fulfilled. For many tasks, especially in robotics, acquiring problem specific databases is not an easy task, specially when considering the additional annotation process. In this context, it would be useful to be able to train models that achieve good performances without requiring annotated real images. 

The success of deep learning applications on autonomous driving and advanced driver assistance systems (ADAS) is unequivocal. For instance, DNNs have been used in scene semantic segmentation~\cite{deeplab2018eccv}, traffic light detection~\cite{behrendt2017icra}, crosswalk classification~\cite{berriel2017cag,berriel2017grsl}, traffic sign detection~\cite{zhu2016cvpr}, pedestrian analysis~\cite{Guidolini2018HandlingPI}, car heading direction estimation~\cite{DBLP:conf/ijcnn/BerrielTCGBSO18} and many other applications. In this work, we focus on the traffic sign detection problem. The goal is to correctly detect (i.e., predict the position on the image) all traffic signs of interest given an image captured by a camera mounted on a vehicle. A traffic sign, in turn, is a specific sign designed to convey information to the road users. It is an important task to be tackled by autonomous driving systems and ADAS, because a traffic sign sets rules which (i) drivers are expected to abide by and (ii) road users rely on while making decisions. Although the complete problem comprises detecting traffic signs and recognizing their category, this work focuses on the first part only, i.e., on finding the position of the signs of interest, since the detection is usually a more complex problem to handle than the classification.

The traffic sign detection and recognition problem has been investigated by the research community for a while. Researchers have been proposing all types of solutions such as the ones using hand-crafted features in model-based solutions~\cite{barnes2008tits},  leveraging simple features in learning-based approaches~\cite{bascon2007tits}, and, the more recent and state-of-the-art, using deep learning based methods~\cite{zhu2016cvpr, garcia2018neurocomputing} that is the focus of this work. A detailed and complete review of these methods can be found in~\cite{gudigar2016mta}.

Apart from the major advances on the topic, there are still many issues requiring further investigation, specially when considering deep learning approaches for detection. Such detectors require: (i) expensive annotation, (ii) real images from the target domain,
and (iii) balanced data sets.
The annotation process is expensive because each traffic sign has to be marked with a bound box, which is more difficult than just setting a class for a classification problem.
Moreover, deep learning based detectors are still known for being data hungry, i.e., for requiring many real images to perform well. Therefore, the acquisition of such real-world images of traffic signs can be troublesome, 
because it requires finding many traffic sign samples along the roads. Since traffic legislation changes from country to country, the traffic signs are not standardized across the world and a new data set has to be created for every country.
Finally, the image acquisition process has to generate a balanced number for each class. This would require collecting many more images to have a minimum balance across the classes because some traffic signs are rarer than others under common driving circumstances.

The research community is well aware of each of the aforementioned issues.
First, there are several tools~\cite{vott2018microsoft, scalabel2018berkeley} that attempt to mitigate the costs of annotating databases for detection tasks. In addition, many people are investigating automatic and semi-automatic techniques to aid the annotation process, some of them including human-in-the-loop~\cite{wang2018cvpr}.
Second, there are some works~\cite{oquab2015cvpr, sangineto2018pami} on weakly supervised object detection that try to leverage the massive amounts of data annotated for classification to perform detection tasks. Moreover, some works~\cite{chen2018aaai, kang2018arxiv} have been investigating few-shot object detection, trying to reduce the need for data collection.
Lastly, data imbalance is a well-known issue and its impact on learning-based methods has been widely investigated even before deep learning~\cite{he2008tkde}. The usual tricks, that provide limited robustness, can be roughly categorized in two techniques: data re-sampling and cost-sensitive learning. In the deep learning context, some works~\cite{huang2016cvpr, wag2017neurips, zhou2018kdd} have investigated the impact of these tricks and how to learn deep representations that take the imbalance into account. Most recently, and more related to this work, Generative Adversarial Networks (GANs) have also been used to perform data augmentation. In the context of traffic sign generation,~\cite{sebastian2018icpr} applies a GAN to enable the conditional generation of traffic signs and~\cite{grigorescu2018icra} learns to generate training samples based on few input images to train a classifier. Nonetheless, there is still need for further investigation and solutions that are able to handle these issues altogether, specially in the object detection context. 

In this context, we hypothesize that it is possible to train a deep traffic sign detector without requiring annotated real images from the domain of interest and, yet, achieve a similar performance to those models trained on manually annotated images from the domain of interest. Therefore, this work proposes a novel effortless synthetic-database generation method that requires only (i) arbitrary natural images (i.e., requires no real image from the domain of interest) and (ii) templates of the traffic signs (i.e., templates synthetically created to illustrate the appearance of the category of a traffic sign). The synthetic database is used to train a country optimized deep traffic sign detector. We argue that, in the context of traffic sign detection, the proposed database generation process handles the three previously mentioned issues altogether facilitating the training of country specific detectors. The proposed approach is evaluated using the German Traffic Sign Detection Benchmark (GTSDB) \cite{Houben-IJCNN-2013} that comprises real images of German traffic signs. Results showed that training the Faster R-CNN detector ~\cite{fasterrcnn} to find German traffic signs with the proposed approach leads to 95.66\% mAP with a low rate of false positives. The proposed method is able to detect traffic signs with an average precision, recall and F1-score of about 94\%, 91\% and 93\%, respectively. The experiments surprisingly show that detectors can be trained with simple data generation methods and without problem domain data for the background, which is in the opposite direction of the common sense for deep learning.
\section{Proposed Method}

The proposed method (illustrated in \Cref{fig:overview}) comprises mainly the generation of a synthetic training data set that requires no real image from the domain of interest. After that, this synthetic data set is used to train a deep traffic sign detector. Finally, the trained deep detector model can be used to infer the position of traffic signs on real images.

\subsection{Training Database Generation}
\label{sec:samples_generation}

The generation of the training database is three-fold. First, templates of the traffic signs of interest are acquired. Then, background images that do not belong to the domain of interest are collected (e.g., random natural images). Lastly, the training samples (i.e., images with annotated traffic signs) are generated.

\begin{figure}[h]
	\centering
	\includegraphics[width=0.8\linewidth]{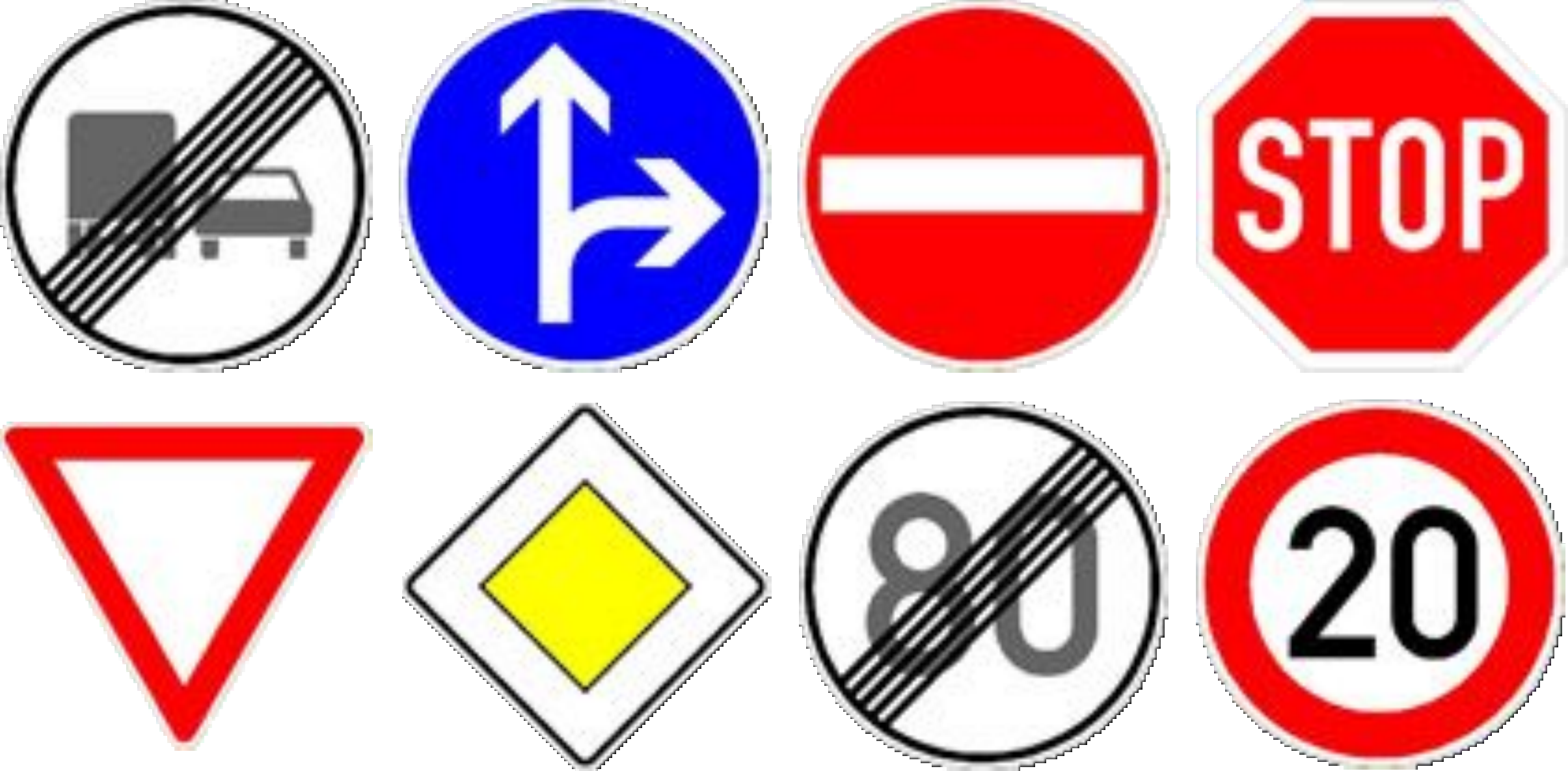}
	\caption{Examples of templates of traffic signs.}
	\label{fig:template_samples}
\end{figure}

\paragraph{Template acquisition}
The first step towards the generation of the training samples is the acquisition of a template for each traffic sign of interest. The traffic signs of interest are those which the system is expected to operate with. Frequently, traffic signs are part of country-wise specific legislations defined by governmental agencies. This usual country-wise standardization helps the acquisition of templates (which is the goal of this step), because their very definitions (i.e., the templates) are part of pieces of legislation commonly available on-line on the websites of these agencies. In fact, templates are graphic representations of these definitions (see some examples in \Cref{fig:template_samples}). In addition, some of the publicly available data sets for traffic sign detection (e.g.,~\cite{Houben-IJCNN-2013}) also distribute the templates of the classes annotated in their samples. All of this makes it easy and convenient to acquire templates virtually for any given set of standardized traffic signs worldwide. These templates are acquired and stored to be used later.

\begin{figure*}[t]
	\centering
	\includegraphics[width=\linewidth]{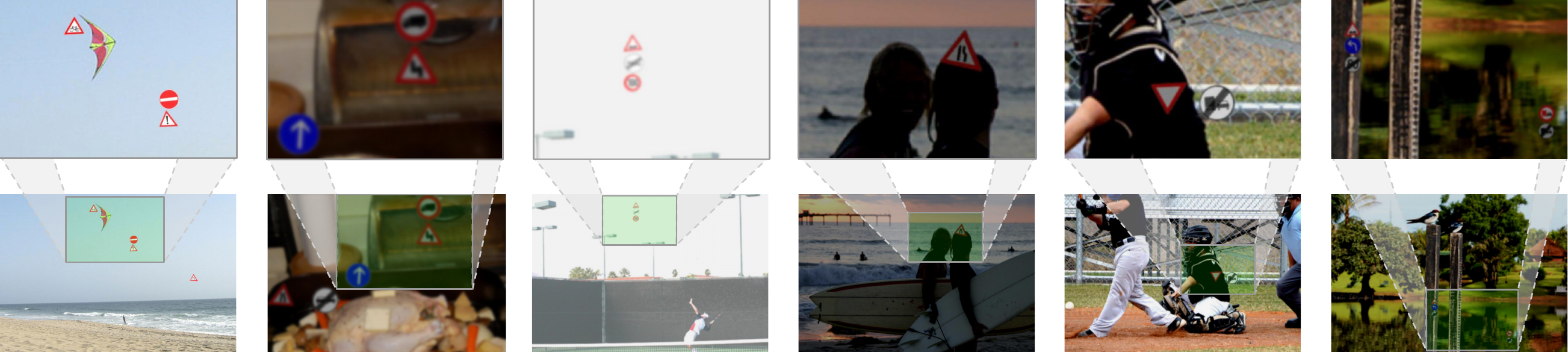}
	\caption{Some training samples can be seen in the bottom row, and zoomed-in figures, highlighting the regions with traffic signs, can be seen in the top row. These samples were generated using the process described in~\Cref{sec:samples_generation}.}
	\label{fig:blend_samples}
\end{figure*}

\paragraph{Background acquisition}
In addition to the templates of the traffic signs of interest, the system requires images to use as background of the training samples. Although the natural choice would be images that belong to the domain of interest, e.g., images of roads, highways, streets, etc.; we argue that this is not required. Not only that, but choosing images from the domain of interest to be used as background may introduce unwanted noise in the training data if not carefully annotated. Images from the domain of interest eventually will present the object of interest and this object will be treated as background as well. On the other hand, by not constraining the background acquisition to images of the domain of interest, many of the freely available large-scale data sets (e.g., ImageNet~\cite{krizhevsky2012imagenet}, Microsoft COCO~\cite{lin2014microsoft}, etc.) can be exploited. For this work, the Microsoft COCO data set was chosen to be used as background, except for the images containing classes that are closely related to the domain of interest in order to avoid the introduction of noise in the training data set. Details are presented in~\Cref{sec:experiments}. After choosing the background images, the training samples can be generated.

\paragraph{Training samples generation}

The last step of the generation of the training database is blending the background images and the templates of the traffic signs to generate the training samples. This blending process aims to reduce the appearance difference between the background and the templates. If successful, the process is advantageous because it may tackle all the three aforementioned issues: (i) the training samples are automatically annotated, since the position of the traffic signs on the image is defined by the method; (ii) a large-scale database can be generated without much cost, given that there are a lot of different possible combinations between the random natural images (i.e., the backgrounds) and the objects of interest (i.e., the transformed traffic sign templates); and (iii) the training data set will not suffer from imbalance, since the method can sample the classes uniformly. The details of the blending process are described below:

Let 
$\mathcal{B} = \{B_{i}\}_{i=1}^{S}$ 
be the background set with $S$ random natural images, in total; 
$\mathcal{C} = \{C_{i}\}_{i=1}^{M}$ 
be the set of classes; and $\mathcal{T} = \{T_i \mid i \in \mathcal{C} \}$ be the set of templates of traffic signs. First, for training a data set with $N$ samples, the training set is defined as $\mathcal{X} = \{X_1, X_2, \cdots, X_N\} \stackrel{iid}{\sim} \mathcal{B}$. The first step is to randomly change the brightness and contrast of the image by randomly adding and multiplying each training sample, i.e., $\alpha_{i} X_i + \beta$, where $\alpha_{i} \sim \text{\textit{U}}(0.75, 1.25)$, $\beta \sim \text{\textit{U}}(-120, 120)$, and $X_i$ is the background image sampled for the $i$-th training sample. The next step is to add a random amount of $|K^{i}|$ templates into the $i$-th sample $X_i$, where $|K^{i}| \in \{1, 2, 3, 4, 5\}$ and $K^{i}_{j} \sim \mathcal{T}$. The first template is positioned at random on the background image. Then, for each template $K^{i}_{j}$, there is a 40\% chance that $K^{i}_{j+1}$ will be placed immediately below it instead of randomly. Also, if $\langle K^{i}_{j}, K^{i}_{j+1} \rangle$ were placed together, there is a 50\% chance that $K^{i}_{j+2}$ will be placed immediately below $K^{i}_{j+1}$. This step attempts to mimic a common behavior in real world (considering the country of interest), where sometimes two or three signs are seen together, one immediately below the other. The process of adding a template into a background image is as follows: (i) multiply the template $K^{i}_{j}$ by the same $\alpha_{i}$ used on the background image $X_i$; (ii) apply geometric transformations, such as random perspective changes,
rotation $\theta \sim \text{\textit{U}}(-10°, 10°)$, and scale (according to the minimum and maximum range of operation desired); (iii) adjust the brightness by adding to the template the average of the region on which the template is being added, minus a constant;
(iv) add noise, i.e., jitter; (v) place the template into a random position (unless it is tied to another template) with no intersection with the others; and (vi) fade the borders of the template to create a smooth transition from the template to the background image. Lastly, a blur $\sigma \sim U(0, 7 \times \textrm{scale})$ is applied to the resulting image, generating the final training sample.
Some training samples can also be seen in the \Cref{fig:blend_samples}.

Finally, it is important to generate the templates according to the range of operation of the application. Therefore, it is important to set up the minimum and maximum size of the templates to be detected and sample the random scales accordingly. This procedure can be seen as a calibration step on which, in a real-world application, one could determine the minimum and maximum size of a traffic sign by looking at few images of a particular target camera. Models and code will be made available at https://github.com/LCAD-UFES/publications-tabelini-ijcnn-2019/blob/master/README.md.

\subsection{Model Training and Inference}
\label{sec:training}
After generating the training database, a deep detector can be trained. In this work, one of the state-of-the-art deep detectors was chosen to be trained, i.e., the Faster R-CNN~\cite{fasterrcnn}. Roughly, the Faster R-CNN is a 2-step detection framework, i.e., (i) the Region Proposal Network (RPN) predicts regions that are likely to contain an object, then (ii) the rest of the framework refines the predicted regions and predicts the class of each object. After trained, the Faster R-CNN can be used to process RGB input images and to predict the bounding boxes, classes, and confidence scores of the predicted traffic signs. It is worth noting that, in this work, there is only the class traffic sign (i.e., there is no distinction between them).

\section{Experimental Methodology}
\label{sec:experiments}

This section introduces the data sets used to train and evaluate the detection models, the metrics for performance
quantification, and the experiments conducted to validate our proposal. The experimental platform is described at the
end of the section.

\subsection{Data sets}

\subsubsection{MS-COCO}

The Microsoft COCO (MS-COCO) \cite{lin2014microsoft} is a large-scale data set (more than 200k labeled images
divided into training and test sets) designed for the tasks of object detection, segmentation, and visual captioning.
For this work, the images of the 2017 version of MS-COCO are used as background for the traffic signs
templates, as described in \Cref{sec:samples_generation}. More specifically, the sign templates are overlaid onto the images of the 
MS-COCO training partition in order to train the traffic sign detector proposed in this work. For our purposes,
traffic-related scenes should be disregarded, which is done by filtering out those images originally labeled as
``traffic light'', ``bicycle'', ``car'', ``motorcycle'', ``bus'', ``truck'', ``fire hydrant'', ``stop sign'',
and ``parking meter''. Images with height less than 600 pixels or width less than 400 pixels are also removed, totaling at the
end 58078 images. The remaining images are further uniformly scaled so that the shortest dimension has 1500 pixels.
Finally, the central 1500 $\times$ 1500 pixels area is cropped from the scaled image. This same procedure is conducted on the MS-COCO
test partition to construct a validation set used in the experiments, as further detailed in \Cref{sec:exp}.

\subsubsection{GTSDB}

The German Traffic Sign Detection Benchmark (GTSDB) \cite{Houben-IJCNN-2013} is publicly available with a total of 900 images (1360 $\times$ 800 pixels), being 600 of them separated for training, and 300 for test. The original data set was filtered in order to keep only those images with signs annotation, resulting in 506 and 235 images for training and test, respectively. Training and evaluation with these images provides the expected upper bound performance for the task of detection traffic signs on the GTSDB benchmark.

The evaluation of the proposed system takes advantage of the sign templates available in GTSDB, which were superimposed onto the MS-COCO images to construct the synthetic training base. In addition, the GTSDB’s test images were used to assess the performance of the proposed method. 
\subsection{Performance Metrics}

In addition to the precision, recall and F1-score metrics, the Mean Average Precision (mAP) was also used to quantify the detection performance. The Average Precision (AP) metric, from which mAP is derived,
follows the same approach in the PASCAL VOC 2012 challenge \cite{pascal-voc-2012}. Basically, AP is defined as the approximate area under
the precision/recall curve obtained for a fixed IoU threshold (0.7, in this work). Then, the mAP value is the
average of APs for all object categories. The mAP value matches the AP's since the categories of the objects (i.e., the traffic signs)
are not taken into consideration.

\subsection{Training the Faster R-CNN}

The training of the Faster R-CNN for both scenarios follows a similar setup. First, off-line data augmentation
was performed. The basic procedure introduced in \Cref{sec:samples_generation} was also adopted in the training of the baselines, except for
the blur and brightness parameters: $\sigma \sim \text{\textit{U}}(0, 2)$ (blur) and $\beta \sim \text{\textit{U}}(-40, 40)$ (brightness). At training time, the input images are resized to have the smaller side with 1500 pixels, a default operation defined by the detection framework. This transformation affects
only the GTSDB (ending up with 1500$\times$882) training data since MS-COCO with templates was already adjusted to this size. Additionally,
on-line augmentation consisting of random vertical and horizontal flipping is applied on the resized images.
The scales and the aspect ratio of the Faster R-CNN anchors were calibrated to be in the range of interest of the application, i.e., with assigned values of \{0.5, 1.0, 2.0, 4.0, 8.0, 16.0\} and  1:1 respectively.
In the training, images are processed in batches of size 1, and the learning rate is initially set to 0.001.
After 20k iterations, the learning rate is decayed to 10\% of its initial value, and the training finishes
after 70k iterations. The rest of the setup (e.g., loss function and optimization algorithm)
follows the default training procedure detailed in the original paper \cite{fasterrcnn}.
It is important to highlight that the particular labels of the templates are disregarded in the training
since the goal is detecting traffic signs without identifying them.

\subsection{Experiments}
\label{sec:exp}

The conducted experiments assess performance on the GTSDB data set by
training the detector on MS-COCO with templates (proposed method) and testing on the GTSDB test set. The results are compared to baselines (lower- and upper-bounds) that are trained on the GTSDB's training set itself and tested on the GTSDB test set. The upper-bound baseline makes use of the entire training set, whereas one training image of each traffic-sign category is used to train the lower-bound baseline. For each scenario, the metrics statistics for 10 training-test runs are recorded. Multiple runs with training images in different presentation order and different seeds are intended to assess robustness and prevent misleading analysis due to non-determinism.
As a conservative measure, the best result was chosen for the baselines among the 70k iterations (out of 7 checkpoints at each 10k iterations). For the proposed method, on the other hand, a validation set comprising synthetic images, i.e., templates onto MS-COCO images, is used to select the best model to be considered. Therefore, the model version that maximizes mAP on the validation set out of 7 checkpoints at each 10k iterations is considered for the evaluation. The process is repeated for each of the 10 trainings.

For mAP analysis, all detections are considered since its computation does not assume any confidence threshold. Precision, recall, and F1-score metrics, on the other hand, require the suppressing of low confidence detections in order to analyze the performance of the deployed detector. To this purpose, the confidence threshold for a trained model was assigned the value that yielded the highest F1-score on the validation set (i.e., MS-COCO test set with templates). This procedure avoids, for instance, favoring recall against precision, which happens when a low threshold is manually chosen.

Finally, we analyze the ratio of recovered objects for each category (this was named category-wise recall). It is important to note that, in this work, the detector does not assign the specific category of the predicted traffic signs. In other words, the confidence value is linked to the object specific category, but with a general class \emph{traffic sign}. Therefore, a traffic sign object (of any category) is said to be recovered correctly if it is intersected by any prediction (with IoU $\geq$ 0.7) with confidence level equal or greater than a threshold (the same defined for precision/recall).

\subsection{Experimental Platform}
The training and evaluation of the models were conducted on two Intel Xeon CPU E5606 (2.13GHz) PCs and a NVIDIA	TITAN Xp GPU with 12 GB, one with 24GB of RAM, and the other with 31 GB. The data set was processed on an Intel Xeon E7-4850 v4 (2.10GHz) PC with 128 vCPU (only 100 were used), 252GB of RAM. For Faster R-CNN, we adopted a	Tensorflow implementation that is publicly available\footnote{https://github.com/endernewton/tf-faster-rcnn}. Training sections take nearly 7 hours for a single run on MS-COCO with templates, and 12 hours on the GTSDB training set.
\section{Results and Discussion}
\label{sec:results}

\begin{figure}[h]
	\centering
	\includegraphics[width=\columnwidth,height=80px]{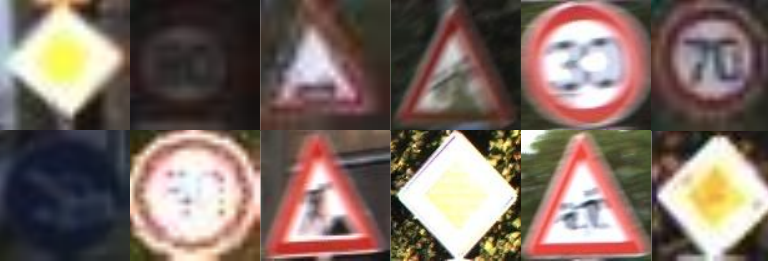}
	\caption{Challenging cases where the proposed method was successful.}
	\label{fig:challenging_cases}
\end{figure}

The mAP of the proposed template-based detector was 95.66 $\pm$ 0.53\%, where these values stand for the mean and standard deviation for the 10 runs, respectively. In other words, a mAP higher than 95\% was achieved without using any image of the application domain. \Cref{fig:challenging_cases} shows some challenging situations our method was able to handle (the best model of the 10 runs was considered here): partial occlusion, physical deterioration, and poor image quality (i.e., blur, noise, low contrast/illumination). Additional qualitative results are available in the video\footnote{https://youtu.be/zK5qaY3uzhE}.

\begin{figure}[h!]
	\centering
	\includegraphics[height=200pt]{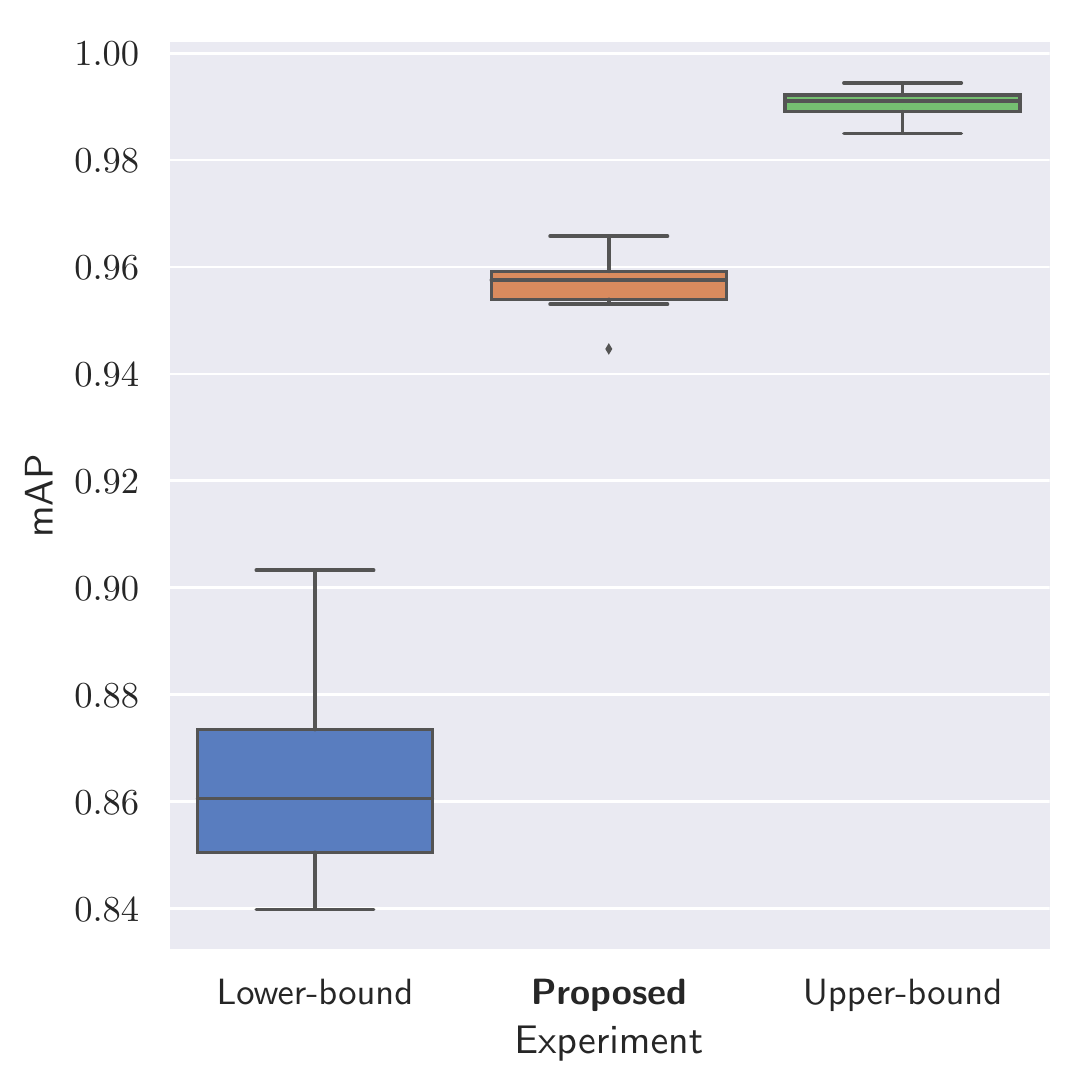}
	\caption{Performance of the models in terms of mAP. The boxplot is the result of 10 runs.}
	\label{fig:comp-map}
\end{figure}

Nevertheless, the resulting mAP should be viewed in light of the baselines' performance (\Cref{fig:comp-map}). The upper-bound mAP was 3.39\% higher, on average, than the proposed method, whereas the lower-bound (i.e., training with one exemplar of each category) was 9.13\% lower. As expected, increasing the collection of annotated traffic scenes yields a better performance, however this implies more human effort. Furthermore, it should be considered that the mAP difference between the upper-bound baseline and the proposed method reflects the fact of the baseline's training and test sets share a particular geographic context (localities near Bochum, Germany). This implies more similarity between the two sets with respect to the overall appearance and the traffic scene structure, which is hard to be reached in real driving applications. By using natural images, the structural dependency is disregarded completely.

\begin{figure}
	
	\centering
	\subfigure[False positives]{\includegraphics[width=\columnwidth,height=80px]{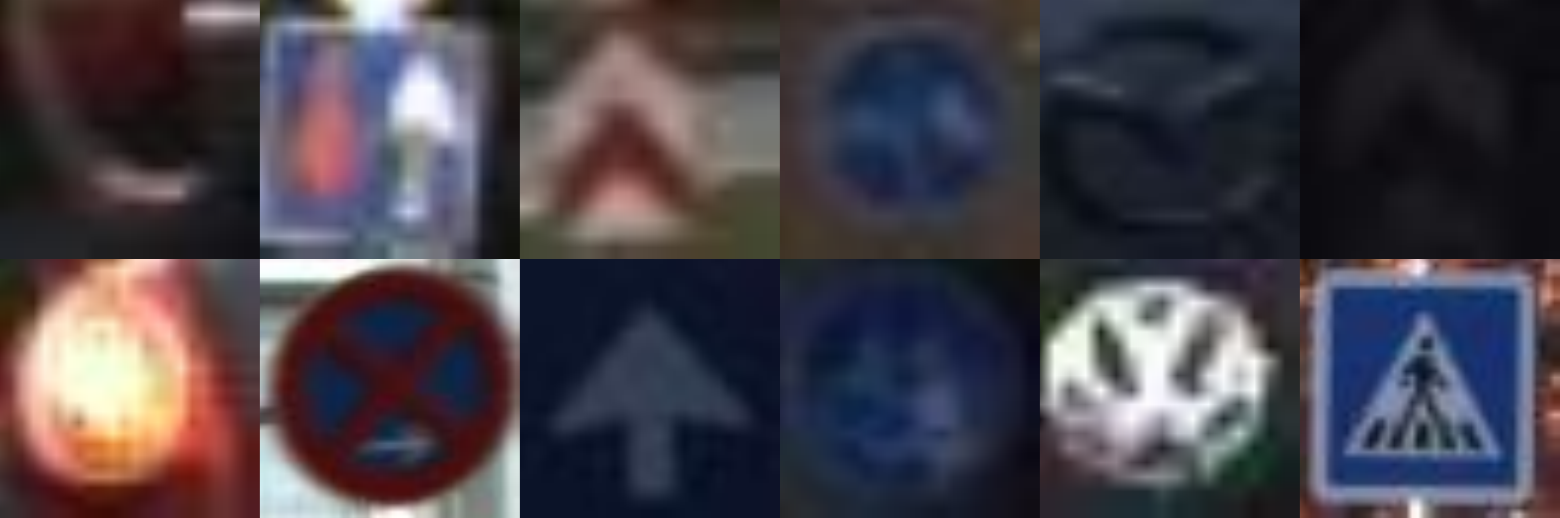}
	\label{fig:pos-samples}}
	
	\qquad
	\subfigure[False negatives]{\includegraphics[width=\columnwidth,height=40px]{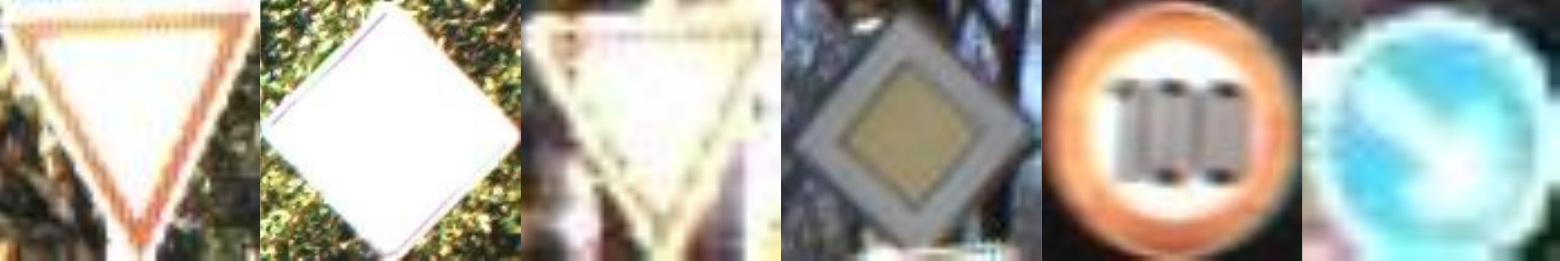}
	\label{fig:neg-samples}}
	\caption{Samples of false positives/negatives produced with the proposed method.}
	\label{fig:samples}
	
\end{figure}

For further analysis, the confidence threshold was fixed according to the validation procedure described in the previous section, and the average precision, recall, and F1-score for the proposed method was measured. The respective values were 90.76\%, 91.99\%, and 0.9135. The precision metric is affected by the amount of false positives it produces. An important observation is that 43.43\% of the false alarms (some samples are shown in \Cref{fig:pos-samples}) indeed resembles to traffic signs, however they were not assigned as objects of interest in the ground-truth annotation. By disregarding these instances through visual inspection of the detected cases, the precision would increase to 94.66\% and, as a consequence, the F1-score would achieve 0.933. The full list of images under this condition is available here\footnote{https://github.com/LCAD-UFES/publications-tabelini-ijcnn-2019/blob/master/False-FalsePositives.pdf}. In addition to these cases, a substantial amount of false positives (50.86\%) include other elements of the driving domain, such as rear-view mirrors, other types of signs, and headlights. We conjecture that this is due to the fact these elements are more scarce in natural images, which hinders learning them as negative samples (i.e., non traffic signs).

\begin{figure}[t]
	\centering
	\includegraphics[width=\columnwidth]{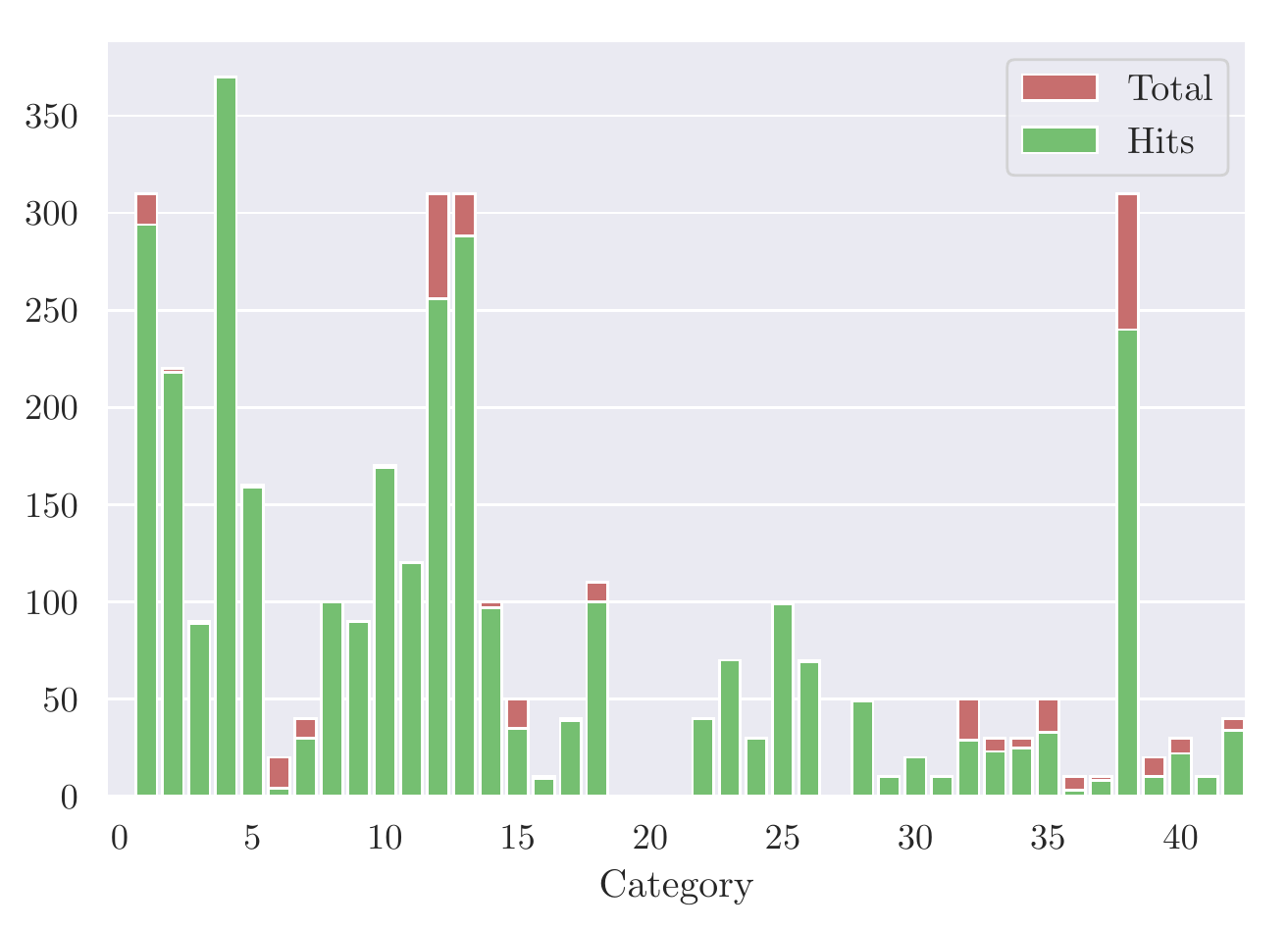}
	\caption{Category-wise recall (non-normalized) accumulated for 10 runs. The red bars represent the number of traffic signs of each category in the test set multiplied by 10, since we account for 10 runs, whereas the green bars denote the quantity of recovered traffic signs accumulated for the 10 runs.}
	\label{fig:recall_per_class}
\end{figure}

For a more detailed evaluation, the category-wise recall performance (\Cref{fig:recall_per_class}) was also analyzed. The red bars in the graph represent the number of traffic signs of each category in the test set multiplied by 10, since we account for 10 runs. The green bars denote the quantity of recovered traffic signs accumulated for the 10 runs (i.e., for the 10 evaluated models). In summary, the proposed method was able to perfectly recover 28.95\% of the 38 categories. For categories with more than 5 exemplars (i.e., red bars reaching 50 in the graph) in the test set, the recall was under 90\% only for the categories 12 and 38. These results show the great potential of the presented method.

Although our method achieved notably good results on the test set, we were also interested to assess its performance on a larger data set. In this context, we evaluated the proposed method on the full GTSDB (i.e., the training and testing sets together). On this data set, the proposed system achieved a mAP of 93.17\%, with average precision, recall and F1-score (the latter three values are according to the best F1-score described previously), of 90.59\%, 89.15\% and 0.8984, respectively. However, this result can not be compared to the baselines, since data from the training set (now included in the test set) is not available anymore.

\section{Conclusion}

Solving challenging tasks with deep neural networks usually requires an annotated data set with real image samples belonging to the context of the problem.
The human effort and other costs involved in gathering such data has motivated research
on alternative ways to train the models. In particular, this work leverages templates
to teach a deep model to detect traffic signs in real traffic scenes. Besides eliminating the need
for real traffic signs, we also propose a more flexible and effortless construction of the training set by superposing the templates on natural images, i.e., arbitrary background images available in computer vision benchmarks.

Experimental results on the German Traffic Sign Detection Benchmark (GTSDB) showed that models trained using automatically-labeled data without images of the problem context achieved, on average, 95.66\% of mAP,
whereas using the GTSDB training set itself (i.e., real annotated data of the problem domain) yielded an improvement of only +3.39\%. In other words,
we can achieve comparable performance to the reference (upper bound) baseline without the need of capturing or using traffic scenes, as well as the need of (human-made) object detection annotation.

The results obtained with this work are very surprising because they show that, for some applications, detection models can be trained out of the context of the problem and still achieve good performance. Moreover, they actually open up doors for new investigations about training of deep architectures. This achievements naturally lead us to inquire about its applicability in other application domains. Future work includes investigating the traffic sign problem with category recognition and a more comprehensive investigation of the tasks and scenarios our proposal could be applied on.

\section*{Acknowledgment}

The authors thank the NVIDIA Corporation for their kind donation of the GPUs used in this research.

\bibliographystyle{IEEEtran}
\bibliography{refs}

\begin{thebibliography}{10}
\providecommand{\url}[1]{#1}
\csname url@samestyle\endcsname
\providecommand{\newblock}{\relax}
\providecommand{\bibinfo}[2]{#2}
\providecommand{\BIBentrySTDinterwordspacing}{\spaceskip=0pt\relax}
\providecommand{\BIBentryALTinterwordstretchfactor}{4}
\providecommand{\BIBentryALTinterwordspacing}{\spaceskip=\fontdimen2\font plus
\BIBentryALTinterwordstretchfactor\fontdimen3\font minus
  \fontdimen4\font\relax}
\providecommand{\BIBforeignlanguage}[2]{{%
\expandafter\ifx\csname l@#1\endcsname\relax
\typeout{** WARNING: IEEEtran.bst: No hyphenation pattern has been}%
\typeout{** loaded for the language `#1'. Using the pattern for}%
\typeout{** the default language instead.}%
\else
\language=\csname l@#1\endcsname
\fi
#2}}
\providecommand{\BIBdecl}{\relax}
\BIBdecl

\bibitem{oursurvey2019arxiv}
C.~Badue, R.~Guidolini, R.~V. Carneiro, P.~Azevedo, V.~B. Cardoso, A.~Forechi,
  L.~F.~R. Jesus, R.~F. Berriel, T.~M. Paixão, F.~Mutz, T.~Oliveira-Santos,
  and A.~F. De~Souza, ``{Self-Driving Cars: A Survey},'' \emph{arXiv preprint
  arXiv:1901.04407}, 2019.

\bibitem{deeplab2018eccv}
L.-C. Chen, Y.~Zhu, G.~Papandreou, F.~Schroff, and H.~Adam, ``{Encoder-Decoder
  with Atrous Separable Convolution for Semantic Image Segmentation},'' in
  \emph{European Conference on Computer Vision (ECCV)}, 2018.

\bibitem{behrendt2017icra}
K.~Behrendt, L.~Novak, and R.~Botros, ``A deep learning approach to traffic
  lights: Detection, tracking, and classification,'' in \emph{International
  Conference on Robotics and Automation (ICRA)}, 2017, pp. 1370--1377.

\bibitem{berriel2017cag}
R.~F. Berriel, F.~S. Rossi, A.~F. de~Souza, and T.~Oliveira-Santos,
  ``{Automatic Large-Scale Data Acquisition via Crowdsourcing for Crosswalk
  Classification: A Deep Learning Approach},'' \emph{Computers \& Graphics},
  vol.~68, pp. 32--42, 2017.

\bibitem{berriel2017grsl}
R.~F. Berriel, A.~T. Lopes, A.~F. de~Souza, and T.~Oliveira-Santos, ``{Deep
  Learning-Based Large-Scale Automatic Satellite Crosswalk Classification},''
  \emph{Geoscience and Remote Sensing Letters}, vol.~14, no.~9, pp. 1513--1517,
  2017.

\bibitem{zhu2016cvpr}
Z.~Zhu, D.~Liang, S.~Zhang, X.~Huang, B.~Li, and S.~Hu, ``{Traffic-Sign
  Detection and Classification in the Wild},'' in \emph{Conference on Computer
  Vision and Pattern Recognition (CVPR)}, 2016.

\bibitem{Guidolini2018HandlingPI}
R.~Guidolini, L.~G. Scart, L.~F. Jesus, V.~B. Cardoso, C.~Badue, and
  T.~Oliveira-Santos, ``{Handling Pedestrians in Crosswalks Using Deep Neural
  Networks in the IARA Autonomous Car},'' in \emph{International Joint
  Conference on Neural Networks (IJCNN)}, 2018.

\bibitem{DBLP:conf/ijcnn/BerrielTCGBSO18}
R.~F. Berriel, L.~T. Torres, V.~B. Cardoso, R.~Guidolini, C.~Badue, A.~F.~D.
  Souza, and T.~Oliveira{-}Santos, ``Heading direction estimation using deep
  learning with automatic large-scale data acquisition,'' in
  \emph{International Joint Conference on Neural Networks, {IJCNN}}, 2018.

\bibitem{barnes2008tits}
N.~Barnes, A.~Zelinsky, and L.~S. Fletcher, ``{Real-Time Speed Sign Detection
  Using the Radial Symmetry Detector},'' \emph{IEEE Transactions on Intelligent
  Transportation Systems}, vol.~9, no.~2, pp. 322--332, 2008.

\bibitem{bascon2007tits}
S.~Maldonado-Bascon, S.~Lafuente-Arroyo, P.~Gil-Jimenez, H.~Gomez-Moreno, and
  F.~Lopez-Ferreras, ``{Road-Sign Detection and Recognition Based on Support
  Vector Machines},'' \emph{IEEE Transactions on Intelligent Transportation
  Systems}, vol.~8, no.~2, pp. 264--278, 2007.

\bibitem{garcia2018neurocomputing}
Álvaro Arcos-García, J.~A. Álvarez García, and L.~M. Soria-Morillo,
  ``{Evaluation of deep neural networks for traffic sign detection systems},''
  \emph{Neurocomputing}, vol. 316, pp. 332--344, 2018.

\bibitem{gudigar2016mta}
A.~Gudigar, S.~Chokkadi, and U.~Raghavendra, ``{A review on automatic detection
  and recognition of traffic sign},'' \emph{Multimedia Tools and Applications},
  vol.~75, no.~1, pp. 333--364, 2016.

\bibitem{vott2018microsoft}
Microsoft, ``{Visual Object Tagging Tool},''
  \url{https://github.com/Microsoft/VoTT}, visited on 12/24/2018.

\bibitem{scalabel2018berkeley}
{Berkeley DeepDrive}, ``{Scalabel},'' \url{https://www.scalabel.ai}, visited on
  12/24/2018.

\bibitem{wang2018cvpr}
K.~Wang, X.~Yan, D.~Zhang, L.~Zhang, and L.~Lin, ``{Towards Human-Machine
  Cooperation: Self-Supervised Sample Mining for Object Detection},'' in
  \emph{Conference on Computer Vision and Pattern Recognition (CVPR)}, 2018.

\bibitem{oquab2015cvpr}
M.~Oquab, L.~Bottou, I.~Laptev, and J.~Sivic, ``{Is Object Localization for
  Free? - Weakly-Supervised Learning With Convolutional Neural Networks},'' in
  \emph{Conference on Computer Vision and Pattern Recognition (CVPR)}, 2015.

\bibitem{sangineto2018pami}
E.~Sangineto, M.~Nabi, D.~Culibrk, and N.~Sebe, ``{Self Paced Deep Learning for
  Weakly Supervised Object Detection},'' \emph{IEEE Transactions on Pattern
  Analysis and Machine Intelligence}, pp. 1--1, 2018.

\bibitem{chen2018aaai}
H.~Chen, Y.~Wang, G.~Wang, and Y.~Qiao, ``{LSTD: A Low-Shot Transfer Detector
  for Object Detection},'' in \emph{AAAI Conference on Artificial
  Intelligence}, 2018.

\bibitem{kang2018arxiv}
B.~Kang, Z.~Liu, X.~Wang, F.~Yu, J.~Feng, and T.~Darrell, ``{Few-shot Object
  Detection via Feature Reweighting},'' \emph{arXiv preprint arXiv:1812.01866},
  2018.

\bibitem{he2008tkde}
H.~He and E.~A. Garcia, ``Learning from imbalanced data,'' \emph{IEEE
  Transactions on Knowledge \& Data Engineering}, no.~9, pp. 1263--1284, 2008.

\bibitem{huang2016cvpr}
C.~Huang, Y.~Li, C.~Change~Loy, and X.~Tang, ``{Learning Deep Representation
  for Imbalanced Classification},'' in \emph{Conference on Computer Vision and
  Pattern Recognition (CVPR)}, 2016.

\bibitem{wag2017neurips}
Y.-X. Wang, D.~Ramanan, and M.~Hebert, ``{Learning to Model the Tail},'' in
  \emph{Advances in Neural Information Processing Systems (NeurIPS)}, 2017, pp.
  7029--7039.

\bibitem{zhou2018kdd}
D.~Zhou, J.~He, H.~Yang, and W.~Fan, ``{SPARC: Self-Paced Network
  Representation for Few-Shot Rare Category Characterization},'' in
  \emph{International Conference on Knowledge Discovery and Data Mining (KDD)},
  2018, pp. 2807--2816.

\bibitem{sebastian2018icpr}
C.~Sebastian, R.~Uittenbogaard, J.~Vijverberg, B.~Boom, and P.~H. de~With,
  ``{Conditional Transfer with Dense Residual Attention: Synthesizing traffic
  signs from street-view imagery},'' in \emph{International Conference on
  Pattern Recognition (ICPR)}, 2018.

\bibitem{grigorescu2018icra}
S.~M. Grigorescu, ``{Generative One-Shot Learning (GOL): A Semi-Parametric
  Approach to One-Shot Learning in Autonomous Vision},'' in \emph{International
  Conference on Robotics and Automation (ICRA)}, 2018, pp. 7127--7134.

\bibitem{Houben-IJCNN-2013}
S.~Houben, J.~Stallkamp, J.~Salmen, M.~Schlipsing, and C.~Igel, ``{Detection of
  Traffic Signs in Real-World Images: The German Traffic Sign Detection
  Benchmark},'' in \emph{International Joint Conference on Neural Networks
  (IJCNN)}, 2013.

\bibitem{fasterrcnn}
S.~Ren, K.~He, R.~Girshick, and J.~Sun, ``{Faster R-CNN: Towards Real-Time
  Object Detection with Region Proposal Networks},'' \emph{IEEE Transactions on
  Pattern Analysis and Machine Intelligence}, vol.~39, no.~6, pp. 1137--1149,
  2017.

\bibitem{krizhevsky2012imagenet}
A.~Krizhevsky, I.~Sutskever, and G.~E. Hinton, ``{ImageNet Classification with
  Deep Convolutional Neural Networks},'' in \emph{Advances in Neural
  Information Processing Systems (NeurIPS)}, 2012, pp. 1097--1105.

\bibitem{lin2014microsoft}
T.-Y. Lin, M.~Maire, S.~Belongie, J.~Hays, P.~Perona, D.~Ramanan,
  P.~Doll{\'a}r, and C.~L. Zitnick, ``{Microsoft COCO: Common Objects in
  Context},'' in \emph{European Conference on Computer Vision (ECCV)}, 2014.

\bibitem{pascal-voc-2012}
M.~Everingham, L.~Van~Gool, C.~K.~I. Williams, J.~Winn, and A.~Zisserman,
  ``{The {PASCAL} {V}isual {O}bject {C}lasses {C}hallenge 2012 {(VOC2012)}
  {R}esults},''
  http://www.pascal-network.org/challenges/VOC/voc2012/workshop/index.html.

\end{thebibliography}

\end{document}